\documentclass[letterpaper, 10 pt, conference]{ieeeconf}  

\IEEEoverridecommandlockouts                              
\usepackage[utf8]{inputenc}
\usepackage[T1]{fontenc}
\usepackage[textsize=scriptsize, disable]{todonotes}
\usepackage{lipsum}
\usepackage{algorithm}
\usepackage{algpseudocode}
\usepackage{url}
\usepackage{blindtext}
\usepackage[separate-uncertainty = true]{siunitx}
\usepackage{color,soul}
\usepackage{amsmath} 
\usepackage{amsfonts}
\usepackage{amssymb}  

\usepackage{shortcuts_ln} 

\algblock{ParFor}{EndParFor}
\algnewcommand\algorithmicparfor{\textbf{for}}
\algnewcommand\algorithmicpardo{\textbf{do parallel}}
\algnewcommand\algorithmicendparfor{\textbf{end\ for}}
\algrenewtext{ParFor}[1]{\algorithmicparfor\ #1\ \algorithmicpardo}
\algrenewtext{EndParFor}{\algorithmicendparfor}

\newcommand\pl[1]{}

\newcommand{\insertref}[1]{\todo[color=green!40]{#1}}

\makeatletter
\newcommand\fs@betterruled{%
  \def\@fs@cfont{\bfseries}\let\@fs@capt\floatc@ruled
  \def\@fs@pre{\vspace*{5pt}\hrule height.8pt depth0pt \kern2pt}%
  \def\@fs@post{\kern2pt\hrule\relax}%
  \def\@fs@mid{\kern2pt\hrule\kern2pt}%
  \let\@fs@iftopcapt\iftrue}
\floatstyle{betterruled}
\restylefloat{algorithm}
\makeatother

\title{\LARGE \bf
Combining Planning, Reasoning and Reinforcement Learning to solve Industrial Robot Tasks
}

\author{Matthias Mayr$^{1}$, Faseeh Ahmad$^{1}$, Konstantinos Chatzilygeroudis$^{2}$, Luigi Nardi$^{1,3}$ and Volker Krueger$^{1}$
	\thanks{$^{1}$Department of Computer Science, Faculty of Engineering (LTH), Lund University, SE~221~00 Lund, Sweden. E-mail: <firstname>.<lastname>@cs.lth.se.
	}%
	\thanks{$^2$Computer Engineering and Informatics Department (CEID), University of Patras, Greece. E-mail: costashatz@upatras.gr.}
	\thanks{$^{3}$
	Department of Computer Science and Electrical Engineering, Stanford University, CA 94305, USA. E-mail: lnardi@stanford.edu.}
}

\begin{document}

\maketitle
\thispagestyle{empty}
\pagestyle{empty}

\begin{abstract}

One of today's goals for industrial robot systems is to allow fast and easy provisioning for new tasks. Skill-based systems that use planning and knowledge representation have long been one possible answer to this. However, especially with contact-rich robot tasks that need careful parameter settings, such reasoning techniques can fall short if the required knowledge not adequately modeled. We show an approach that provides a combination of task-level planning and reasoning with targeted learning of skill parameters for a task at hand. Starting from a task goal formulated in \textit{PDDL}, the learnable parameters in the plan are identified and an operator can choose reward functions and parameters for the learning process. A tight integration with a knowledge framework allows to form a prior for learning and the usage of multi-objective Bayesian optimization eases to balance aspects such as safety and task performance that can often affect each other.
We demonstrate the efficacy and versatility of our approach by learning skill parameters for two different contact-rich tasks and show their successful execution on a real 7-DOF \emph{KUKA-iiwa}.
\end{abstract}

\section{Introduction}
One of the current trends in industrial manufacturing is to move to smaller batch sizes and higher flexibility of workstations. However, robot programs are often static and cannot be adapted quickly\insertref{CASE paper}. One way to maintain a high flexibility as well as known performance is to use systems based on parameterized \textit{skills}~\cite{krueger19rcim,krueger16ieee, bogh2012does}. These encapsulated abilities typically realize a semantically defined action such as pushing or picking an object. State-of-the-art skill-based software architecture can utilize these skill definitions to create plans for a given task. This allows fast and easy reconfiguration of production lines.

However, when working with contact-rich tasks, it is often difficult to reason about skill parameters. One example are the parameters of a peg insertion search strategy where material properties (e.g. friction) and the robot behavior need to be considered. While it is often possible to create a reasoner that follows a set of rules to determine such skill parameters, it is challenging to implement and to maintain.

Alternatively, it is possible to have operators manually specify and try values for these skill parameters. However, this is a manual process and can be cumbersome.

Finally, it is possible to allow the system to learn by interacting with the environment. Many policy formulations that allow learning (e.g. artificial neural networks) have deficiencies which make their application in an industrial domain challenging. Primarily during the learning phase, dangerous behaviors can be produced and even state-of-the-art RL methods need hundreds of hours of interaction time~\cite{chatzilygeroudis2019survey}. Learning in simulation can help to reduce downtime and dangers for the real system. But still many policy formulations are black boxes for operators and it can be hard to predict their behavior, which could hinder the trust to the system~\cite{edmonds2019tale}. Moreover, the simulation-to-reality gap~\cite{koos2012transferability,mouret201720} is bigger in lower-level control states (i.e. torques), and policies working directly on raw control states struggle to transfer learned behaviors to the real systems~\cite{chatzilygeroudis2019survey}. To address these points, our policy formulation consists of behavior trees (BT) with a motion generator~\cite{rovida182iicirsi}. It has shown to be able to learn interpretable as well as robust behaviors~\cite{mayr21iros} and allows for an easy formulation of parameter priors~\cite{mayr22priors} to accelerate learning.

The formulation of a learning problem for a given task is often not easy and becomes more challenging if factors such as safety or impact on the workstation environment need to be considered.

Therefore, we propose to use multi-objective optimization techniques allow to specify multiple objectives and optimize for them concurrently. This enables operators to
 select from solutions that are optimal for a certain trade-off between the objectives (usually represented as a set of Pareto-optimal solutions).

\section{Related Work}

\subsection{Planning and Learning}

Symbolic planning is combined with learning in \cite{grounds2005combining,gordon2019should,yang2018peorl,sarathy2020spotter}. In~\cite{grounds2005combining}, the PLANQ-Learning algorithm uses a symbolic planner to shape the reward function to be used by the Q-learner to get an optimal policy with on the grid domain. \cite{gordon2019should} uses the combined symbolic planner with reinforcement learning (RL) in a hierarchical framework to solve complex visual interactive question answering tasks. PEORL~\cite{yang2018peorl} integrates symbolic planning and  hierarchical RL (HRL) in the taxi domain and grid world. SPOTTER~\cite{sarathy2020spotter} uses RL to allow the planning agent to discover the new operators required to complete tasks in grid world. In contrast to all these approaches, our approach aims towards real-life robotic tasks in an Industry 4.0 setting.

In~\cite{styrud2021combining},  the authors combine symbolic planning with behavior trees (BT) to solve blocks world tasks with a robot manipulator. They use modified genetic programming~\cite{koza1992genetic} to learn the structure of the BT. In our approach, we focus on learning the parameters of the skills in the BT and utilize a symbolic planner to obtain the structure of the BT.

\section{Approach}

Our approach consists of two main components that interact in different stages of the learning pipeline: First, \textit{SkiROS}~\cite{rovida2017skiros}, a skill-based framework for ROS, which represents the implemented skills with BTs, hosts the world model (digital twin), and interacts with the planner. \textit{SkiROS} is also used to execute behavior trees (BT) while learning and evaluating policies. Second, our learning framework \textit{SkiREIL} that provides the simulation, the integration with the policy optimizer as well as the reward function definition and calculation.

\begin{figure}[tpb]
	{
		\setlength{\fboxrule}{0pt}
		\framebox{\parbox{3in}{
		\includegraphics[width=0.92\columnwidth]{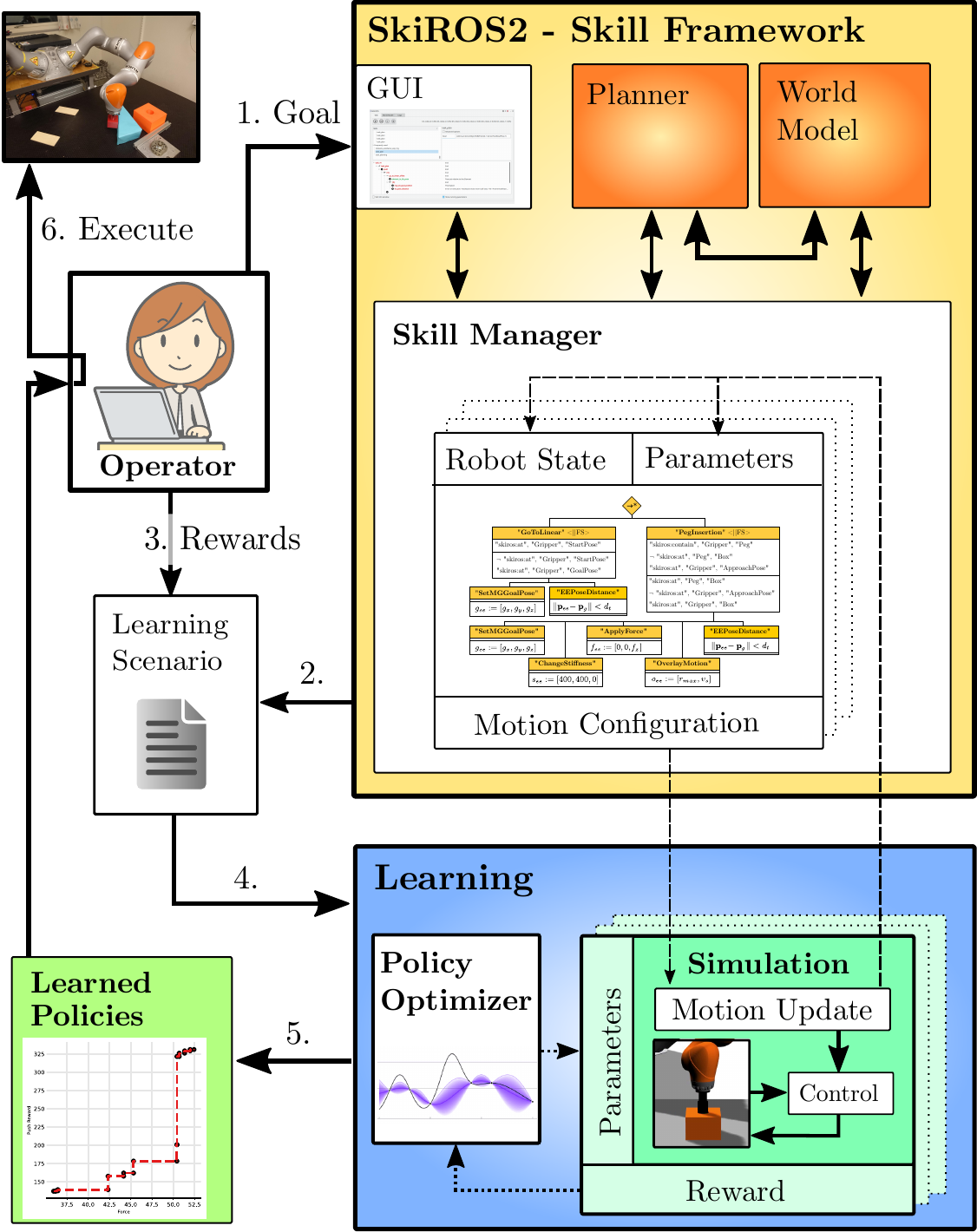}}}
	}
	\caption{The architecture of the system that depicts the pipeline: (1) The operator enters the goal state; (2) a learning scenario for the plan is created; (3) rewards and parameters are specified; (4) learning is conducted using the skills and the information in the world model; (5) after policy learning, the operator can choose which policies to execute on the real system (6).}
	\label{fig:system}
	\vspace{-1em}
\end{figure}
The architecture of the system and the workflow is shown in Figure~\ref{fig:system}: (1) an operator enters the task goal into a GUI; (2) a plan with the respective learning scenario configuration is generated; (3) an operator complements the scenario with objectives and reward functions; (4) learning is conducted in simulation using the skills and information from the world model; (5) in the multi-objective optimization case, a set of Pareto-optimal solutions is generated and presented to the operator; finally, (6) the operator can select a good solution from this set given the desired trade-off between key performance indicators and execute it on the real system.

\subsection{Planning and Knowledge Integration}

The Planning Domain Definition Language (PDDL)~\cite{fox2003pddl2,crosby17icaps} is used to formulate the planning problem. 
We utilize the \textit{SkiROS}~\cite{rovida2017extended} framework that
automatically translates a task into a PDDL planning problem by generating domain description and problem instance using the world model.
We then use the semantic world model (WM) from \textit{SkiROS}~\cite{rovida2017skiros} as the knowledge integration framework. 
Actions and fluents are obtained by utilizing the predicates that have pre- or post-conditions in the world model.
For the problem instance, the objects (robots, arms, grippers, boxes, poses, etc.) in the scene and their initial states (as far as they are known) are used.

After getting the necessary domain description and the problem instance \textit{SkiROS} calls the planner. The goal of the planner is to return a sequence of skills that can achieve the goal conditions of the task. The individual skills are partially parameterized with explicit data from the WM by reasoning about the preconditions. The WM is aware of the skill parameters that need to be learned for the task at hand and they are automatically identified in the skill sequence.

\subsection{Bayesian Optimization} 
\label{sec:bo}
We utilize Bayesian optimization (BO)~\cite{frazier2018tutorial} to learn the skill parameters in a statistically efficient way~\cite{brochu2010tutorial}.
We consider the problem of finding a global minimizer (or maximizer) of an unknown (black-box) objective function $f$:
$\mathbf{s^*} \in \argmin_{\mathbf{s} \in \bbS}
f(\mathbf{s}),$ 
where $\bbS$ is some input design space of interest in $D$ dimensions.
The problem addressed in this paper is the optimization of a 
(possibly noisy) function $f:\bbS \rightarrow \bbR$ 
with lower and upper bounds on the problem variables.
The variables defining $\bbS$ can be real (continuous), integer, ordinal, and categorical as in~\cite{nardi18hypermapper}.
and the derivatives of $f$ are assumed not to be available.
The function $f$ is called black box because we cannot access other information than the output $y$ given an input value $\mathbf{s}$. 
 
BO approximates $\mathbf{s}^*$ with a sequence of evaluations, $y_{1}, y_{2},\ldots, y_t$ at  $\mathbf{s}_1, \mathbf{s}_2, \ldots, \mathbf{s}_t \in \bbS$, which maximizes an utility metric, with each new $\mathbf{s}_{t+1}$ depending on the previous function values. BO achieves this by building a probabilistic surrogate model on $f$ based on the set of evaluated points $\{(\mathbf{s}_i, y_i)\}_{i=1}^{t}$. At each iteration, a new point is selected and evaluated based on the surrogate model which is then updated to include the new point $(\mathbf{s}_{t+1}, y_{t+1})$. 
BO defines an utility metric called the acquisition function, which gives a score to each $\mathbf{s} \in \bbS$ by balancing the predicted value and the uncertainty of the prediction for $\mathbf{s}$. 
We use the implementation of BO found in \textit{HyperMapper}~\cite{nardi18hypermapper}.

\subsection{Multi-objective Optimization}
\label{sec:multi-objective}
Let us consider a multiple objectives minimization (or maximization) over $\bbS$ in $D$ dimensions.
We define $f:\bbS \rightarrow \mathbb{R}^{p}$ as our vector of objective functions $f = (f_{1}, \dots, f_{p})$,
taking $\mathbf{s}$ as input, and evaluating $y = f(\mathbf{s}) + \epsilon$, where $\epsilon$ is a Gaussian noise term. Our goal is to identify the Pareto frontier of $f$, that is,
the set $\Gamma \subseteq \bbS$ of points which are not dominated by any other point,
\ie the maximally desirable $\mathbf{s}$ which cannot be optimized further for any single objective without making a trade-off.
Formally, we consider the partial order in $\mathbb{R}^{p}$: $y \prec y'$ iff $\forall i \in [p], y_{i} \leqslant y'_{i}$
and $\exists j, \, y_{j} \! < \! y'_{j}$, and define the induced order on $\bbS$: $\mathbf{s} \prec \mathbf{s}'$ iff $f(\mathbf{s}) \prec f(\mathbf{s}')$.
The set of minimal points in this order is the Pareto-optimal set $\Gamma = \{\mathbf{s} \in \bbS : \nexists \mathbf{s}'$ such that $\mathbf{s}' \prec \mathbf{s}\}$. We aim to identify $\Gamma$ with the fewest possible function evaluations using BO. 

\section{Experiments}
\label{sec:experiments}
In order to solve a task, the planner determines a sequence of existing pre-defined skills that can achieve the goal condition of the task. This skill sequence is also automatically parameterized to the extend possible, e.g. the goal pose of a movement.
We evaluate our system in two contact-rich scenarios: A) pushing an object with uneven weight distribution to a goal pose and B) inserting a peg in a hole with a \SI{2.5}{\milli\meter} larger radius. We do not have access for reasoners to determine the parameters for these contact-rich tasks, therefore the pure planning-based solutions have a poor performance in reality.

The robot arm used for the physical evaluation is a 7-degree-of-freedom (DOF) \textit{KUKA iiwa} arm controlled by a Cartesian impedance controller.

\subsection{Reward Functions}
\label{sec:rewards}
For each task, we utilize a set of reward functions parameterized that are configured with a weight and are assigned to an objective. In the experiments for this work we used these reward functions:

\begin{enumerate}
    \item \textbf{Task completion}: Assigned when the BT returns success upon task completion. 
    \item \textbf{End-effector distance to a box}: Based on the distance of the end effector to a box
    \item \textbf{Applied wrench}: Cumulative forces applied by the end effector
    \item \textbf{End-effector distance to a goal}: Based on the distance between the end-effector's current pose and a goal pose
    \item \textbf{End-effector-reference-position distance}: Uses the distance between the end effector's reference pose and its current pose
    \item \textbf{Object-pose divergence}: Translational and angular distance between the object's goal pose and its current pose.
\end{enumerate}

\subsection{Peg-in-Hole Task}
\label{sec:peg-in-hole-task}
The BT that is generated uses two skills: 1) $GoToLinear$ skill and 2) $PegInsertion$ skill. The first skill moves the end effector from its current location to the \emph{approach pose} of the hole. Once it is reached, the peg insertion procedure starts. We learned 10 policies for 400 iterations in simulation. To increase robustness of the solutions, we evaluate each parameter configuration in 7 worlds with varying attributes. In this experiment we vary the location of the box and sample one out of 5 start configurations of the robot arm.

\begin{figure}[tpb]
	{
		\setlength{\fboxrule}{0pt}
		\framebox{\parbox{3in}{
			\includegraphics[width=0.95\columnwidth]{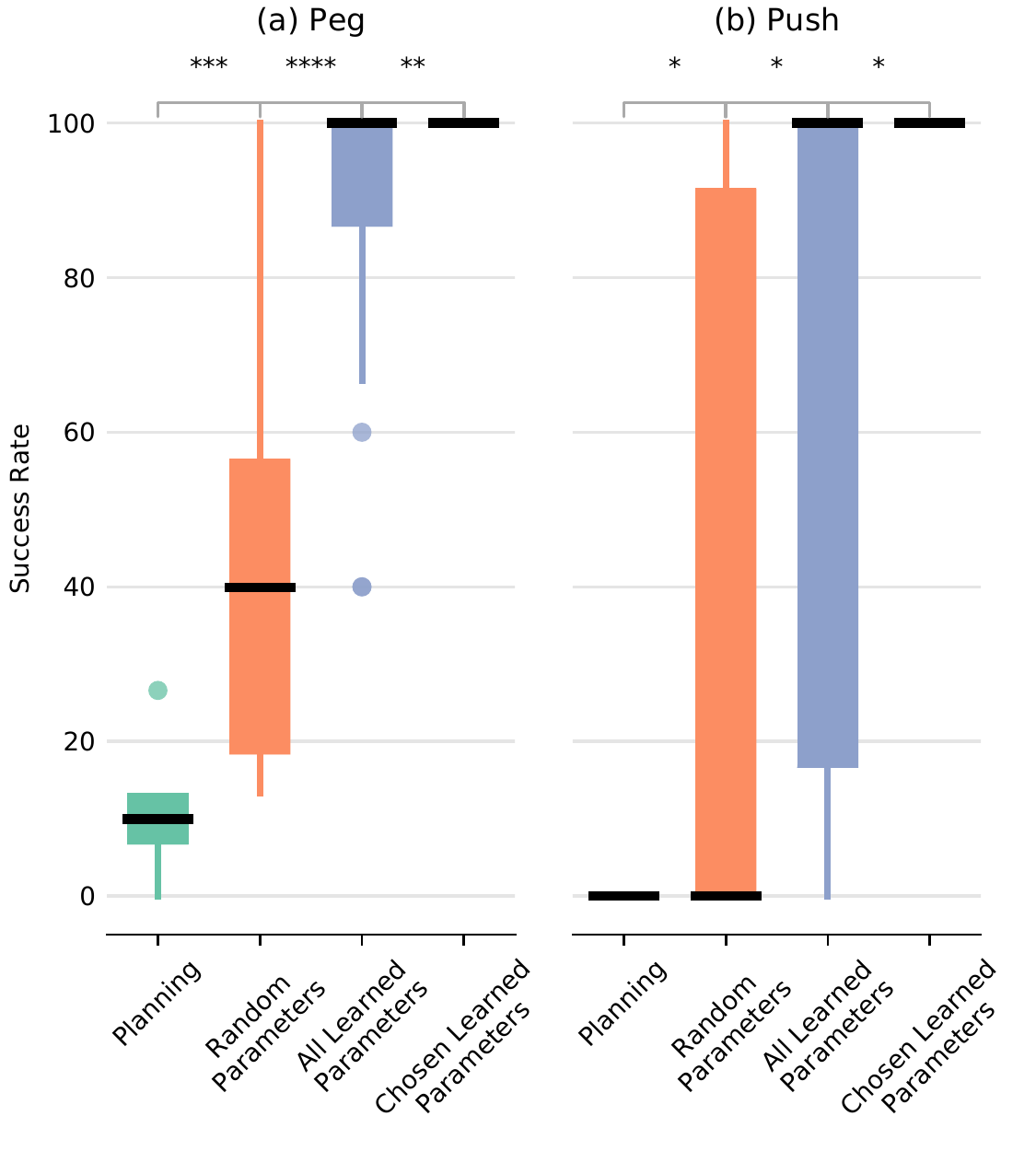}}}
	}
	\caption{The success rates of both experiments. The box plots show the median (black line) and interquartile range (25\(^{th}\) and 75\(^{th}\) percentile); the lines extend to the most extreme data points not considered outliers, and outliers are plotted individually. 
 }
	\label{fig:success-results}
	\vspace{-1.5em}
\end{figure}

The peg insertion task is modeled as a multi-objective task and there are two objectives: 1) successful insertion and 2) applied force.
There are three learnable parameters in this task: 1) downward force applied by the robot arm, 2) radius of the overlay search motion and 3) path velocity of the overlay search motion.

We compare the performance of the learned policies with (1) the outcome of the planner without a parameterized search motion and (2) randomly chosen parameter configurations from the parameter search space used for learning.

The results of a multi-objective optimization are parameters found along a Pareto front. It consist of \num{6.1} points on average. For the results shown in Fig.~\ref{fig:success-results}a, we evaluated the insertion success using the 5 known and additional 10 unknown start configurations of the robot.

\subsection{Push Task}
\label{sec:push-task}

The generated plan for this experiment consists of two skills: 1) $GoToLinear$ skill and 2) $Push$ skill. The first skill moves the end effector from its current location to the \emph{approach pose} of the object.  The push skill then moves the end effector to the object's geometric centre with an optional (learnable) offset in the horizontal (\(x\)) and (\(y\)) directions and pushes in a straight line to the known target location with a learnable offset in the horizontal (\(x\)) and (\(y\)) directions.

The push task is formulated as a multi-objective task. It also has two objectives: 1) task performance and 2) applied force. The first objective is assessed with the pose difference of the object. The second objective accumulates the Cartesian distance between the end-effector reference pose and the actual end-effector pose as a measure of the force that is applied by the controller.

We learn for 400 iterations and repeat the experiment 10 times. We apply domain randomization and each parameter set is evaluated in 7 worlds. Each execution uniformly samples one out of the four start positions for the robot arm as well as varies the location of the object and the goal in the horizontal $(x)$ and $(y)$ directions by sampling from a Gaussian distribution.

We compare the learned solutions with (1) the outcome of a direct planner solution without any offset on the start and goal pose while pushing and (2) 10 sets of random parameters from the search space. We evaluated on the four start configurations used for learning as well as on two additional unknown ones.
The Pareto front contained \num{8.3} points on average, of which some minimize the impact on the environment to an extent that the push is not successful (see Fig.~\ref{fig:success-results}b). An operator can choose a solution that is a good compromise between the success of the task on the real system and the force applied on the environment.

\section{Conclusion}
We have shown a method for effectively combining task-level planning with learning to solve industrial contact-rich tasks. Our method leverages prior information and planning to acquire \emph{explicit} knowledge about the task, whereas it utilizes learning to capture the \emph{tacit} knowledge, i.e., the knowledge that is hard to formalize and which can only be captured through actual interaction.
We formulate a multi-objective optimization scheme so that (1) we handle conflicting rewards adequately, and (2) an operator can choose a policy from the Pareto front and thus actively participate in the learning process.

We evaluated our method on two scenarios using a real \textit{KUKA} 7-DOF manipulator: (a) a peg insertion task, and (b) a pushing task. Both tasks are contact-rich and na\"{i}ve planning fails to solve them and the approach was able to outperform the baseline.

For future work we are looking into multi-fidelity learning and a generalization to different variations of a task~\cite{ahmad2022generalizing}.

\addtolength{\textheight}{-0.1cm}   
\section*{Appendix}
The implementation and a supplemental video are available at:\\\small{\url{https://sites.google.com/ulund.org/SkiREIL}}\\

This work was partially supported by the Wallenberg AI, Autonomous Systems and Software Program (WASP) funded by Knut and Alice Wallenberg Foundation.
This research was also supported in part by affiliate members and other supporters of the Stanford DAWN project—Ant Financial, Facebook, Google, InfoSys, Teradata, NEC, and VMware.


\bibliography{root}
\bibliographystyle{bib/IEEEtran}

\end{document}